\documentclass{article}

\usepackage[preprint]{corl_2020} 
\usepackage{graphicx,caption}
\usepackage{wrapfig}
\usepackage{amsmath}
\usepackage{lipsum}
\usepackage{xcolor}
\usepackage[shortlabels]{enumitem}

\usepackage[acronym]{glossaries}
\makeglossaries

\newacronym{sr}{all-at-once}{standard rollout}
\newacronym{mr}{1-at-a-time}{one-agent-at-a-time rollout}
\newacronym{oomr}{order-optimized}{order-optimized rollout}
\newacronym{mr_api}{1-at-a-time API}{one-agent-at-a-time approximate policy iteration}
\newacronym{amr}{AMR}{approximate multiagent rollout}
\newacronym{amr_b}{AMR-B}{approximate multiagent rollout with base policy signaling and shared belief}
\newacronym{amr_n}{AMR-N}{approximate multiagent rollout with neural network policy signaling and shared belief}
\newacronym{amr_pi}{AMR-PI}{approximate multiagent rollout with PI policy signaling and shared belief}
\newacronym{amr_lc}{AMR-LC}{approximate multiagent rollout with local communication and shared belief}
\newacronym{amr_ilc}{AMR-ILC}{approximate multiagent rollout with intermittent and local communication and shared belief}
\newacronym{amr_ib0}{AMR-IB0}{approximate multiagent rollout with intermittent communication and base policy}
\newacronym{amr_ib1}{AMR-IB1}{approximate multiagent rollout with intermittent communication and base policy signaling}
\newacronym{amr_ibm}{AMR-IBm}{approximate multiagent rollout with intermittent communication and Base policy signaling with m optimization per agent}

\newacronym{amr_in0}{AMR-IN0}{approximate multiagent rollout with intermittent communication and neural network policy}
\newacronym{amr_in1}{AMR-IN1}{approximate multiagent rollout with intermittent communication and neural network policy signaling}
\newacronym{amr_inm}{AMR-INm}{approximate multiagent rollout with intermittent communication and neural network policy signaling with m optimization per agent}

\newcommand{\SB}[1]{\textcolor{black}{#1}}
\newcommand{\SG}[1]{\textcolor{black}{#1}}
\newcommand{\Doubt}[1]{\textcolor{black}{#1}}

\def\a{\alpha}

\def\tl{\tilde}

\title{Multiagent Rollout and Policy Iteration for POMDP with Application to Multi-Robot Repair Problems}

\author{
  Sushmita Bhattacharya\\
  Harvard University\\
  \texttt{sushmita\_bhattacharya@g.harvard.edu} \\
  \And
  Siva Kailas \\
  Arizona State University\\ 
  \texttt{skailas@asu.edu} \\
  \AND
  Sahil Badyal \\
  Arizona State University\\ 
  \texttt{sbadyal@asu.edu} \\
  \And
  Stephanie Gil \\
  Assistant Professor\\
  Harvard University\\
  \texttt{sgil@seas.harvard.edu} \\
  \And
  Dimitri Bertsekas \\
  Fulton Professor\\
  Arizona State University \\
  \texttt{dimitrib@mit.edu} \\
}

\begin{document}
\maketitle
\nolinenumbers

\vspace{-25pt}
\begin{abstract}
    In this paper we consider  infinite horizon discounted dynamic programming problems with finite state and control spaces, partial state observations, and a multiagent structure. We discuss and compare algorithms that simultaneously or sequentially optimize the agents' controls by using multistep lookahead, truncated rollout with a known base policy, and a terminal cost function approximation. 
	Our methods specifically address the computational challenges of partially observable multiagent problems. In particular: 1) We consider rollout algorithms that dramatically reduce required computation while preserving the key cost improvement property of the standard rollout method. The per-step computational requirements for our methods are on the order of $O(Cm)$ as compared with $O(C^m)$ for standard rollout, where $C$ is the maximum cardinality of the constraint set for the control component of each agent, and $m$ is the number of agents. 2) We show that our methods can be applied to challenging problems with a graph structure, including a class of robot repair problems whereby multiple robots collaboratively inspect and repair a system under partial information. 3) We provide a simulation study that compares our methods with existing methods, and demonstrate that our methods can handle larger and more complex partially observable multiagent problems (state space size $10^{37}$ and control space size $10^{7}$, respectively). 
	In particular, we verify experimentally that our multiagent rollout methods perform nearly as well as standard rollout for problems with few agents, and produce satisfactory policies for problems with a larger number of agents that are intractable by standard rollout and other state of the art methods.
	Finally, we incorporate our multiagent rollout algorithms as building blocks in an approximate policy iteration scheme, where successive rollout policies are approximated by using neural network classifiers. While this scheme requires a strictly off-line implementation, it works well in our computational experiments and produces additional significant performance improvement over the single online rollout iteration method.
\end{abstract}
\vspace{-5pt}

\vspace{-5pt}
\section{Introduction}
	\vspace{-5pt}
    We consider the classical partial observation Markovian decision problem (POMDP) with a finite number of states and controls, and discounted additive cost over an infinite horizon. We focus on a version of the problem that has a multiagent character: it involves a control that has multiple components, each corresponding to a different agent. An optimal solution by dynamic programming (DP) is typically intractable. In this paper, we instead propose a suboptimal solution/reinforcement learning approach, whose principal characteristic is the proper exploitation of the multiagent structure to dramatically reduce the computational requirements of the solution method. 
    It is based on the multiagent rollout and policy iteration ideas first proposed in the paper ~\citep{bertsekas2019multiagent} and discussed at length in the research monograph~\citep{Ber20}. 
    \SG{Important distinctions of the current work include explicit treatment of the partial state observability case, and the development of a multiagent decision making framework within the partially observable context.}
    
    \begin{wrapfigure}[11]{R}{0.45\textwidth}
	\begin{center}
		\includegraphics[width=\linewidth]{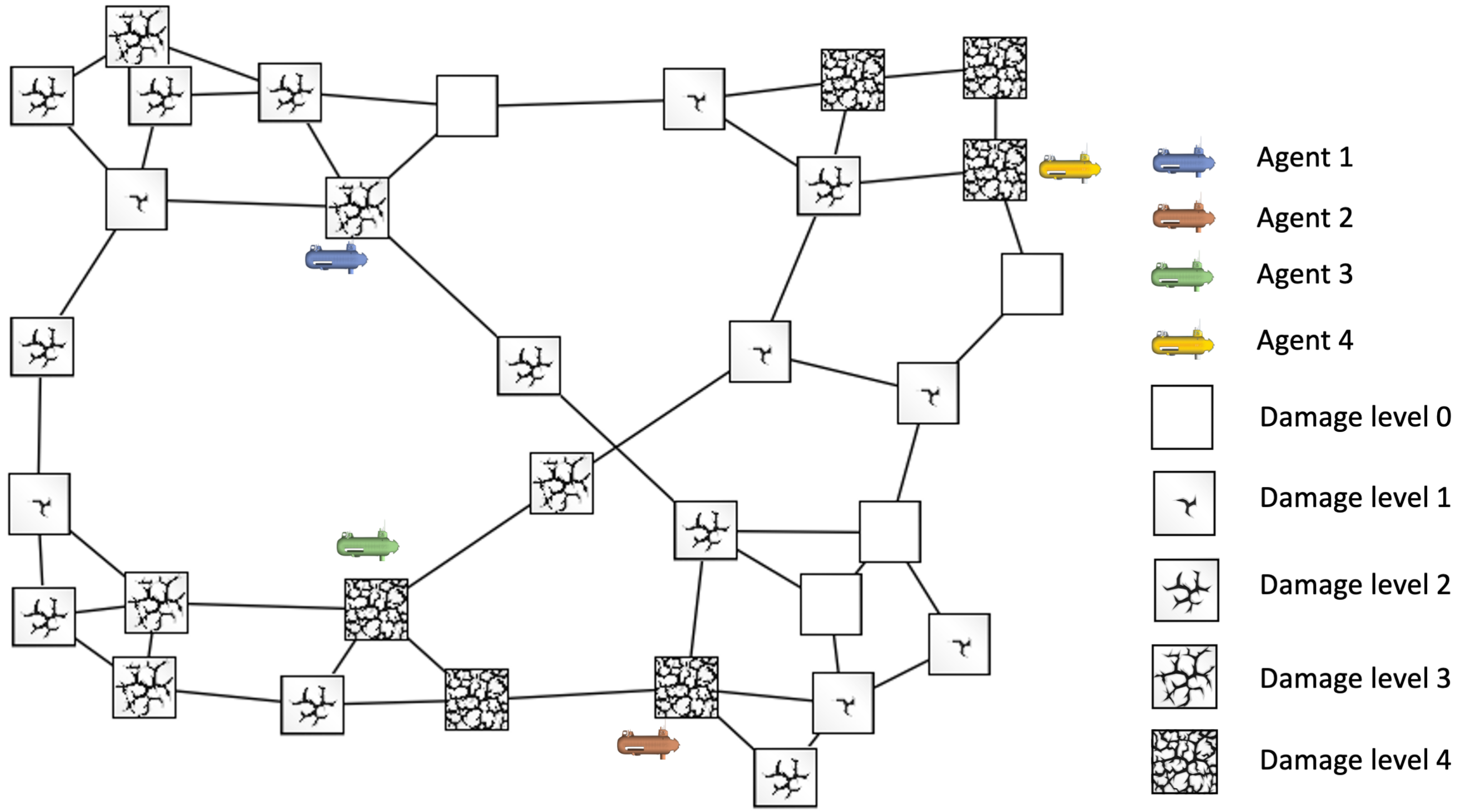}
	\end{center}
	\vspace{-9pt}
	\caption{\small A 2D repair network with 32 nodes and 4 robots.}

	\label{fig:graph}
	\vspace{-10pt}
\end{wrapfigure}

The standard form of rollout, as described in several sources (e.g., the reinforcement learning book~\citep{Ber19}), starts with some easily implementable policy, called the {\it base policy\/}, and produces another policy, the {\it rollout policy\/}, using one-step or multistep lookahead optimization. Its key property is policy improvement: the rollout policy has improved performance over the base policy for all initial states. This is referred to as \emph{the policy improvement property}.

In a multiagent setting, the lookahead optimization portion of the standard rollout algorithm becomes very time-consuming. By contrast, in our multiagent rollout approach, the implementation complexity of the lookahead optimization is dramatically reduced, while maintaining the fundamental policy improvement property. Our multiagent rollout policy may also be implemented approximately by using truncated rollout, and a terminal cost function approximation. In this case, the policy improvement property applies in an approximate form, which is quantified by an error bound that is no worse than the one for the corresponding standard rollout policy. \SG{We assume perfect communication of controls and/or belief states amongst agents. In the appendix of this paper (Section~\ref{sec:communication_less} and~\ref{sec:communication}), we also consider an extension where agents cannot communicate their belief states and controls to one another at all times. 
We note that in the imperfect communication case the cost improvement property does not hold, but we develop and present extensive numerical results comparing the performance of the perfect versus various imperfect communication architectures in the context of multiagent rollout.}

We explore the implementation of our multiagent rollout methods for a class of challenging multi-robot repair problems. Here, it is important to use a policy that can identify and execute critical repairs in minimum time by leveraging coordination among the agents. We demonstrate the power of our method by applying it to a complex repair problem, involving a graph of 32 potentially damaged locations and as many as 10 repair robots/agents (see Fig.~\ref{fig:graph}). This is a number of agents that are well beyond the capabilities of standard POMDP methods. In particular, we present favorable comparisons of our proposed method with the state-of-art software POMCP~\citep{SiV10} and MADDPG~\citep{lowe2017multi} that also provide an approximate solution for POMDP problems.

\vspace{-10pt}
\section{Related Work}
\vspace{-10pt}
Several reinforcement learning algorithms for POMDP problems have been proposed in the literature. 
In particular,~\citep{KLC98} describes a general solution method for POMDP,~\citep{MPK99} discusses a policy search method using finite state controllers, ~\citep{ZhH01, YuB04, Ber19} discuss aggregation-based methods, 
 and ~\citep{BaB01, Yu05, ELP12} consider actor-critic based policy gradient methods.
These methods are fundamentally different from our proposed rollout-based methodologies, as they do not directly rely on cost improvement starting from a base policy.

Among works in the POMDP literature, there are some that like our rollout-based methods, use lookahead minimization, and also try to trade off the length of simulated trajectories with variable length lookahead and pruning. 
In particular, POMCP~\citep{SiV10} uses multistep lookahead and Monte-Carlo Tree Search (MCTS) to generate a suboptimal policy, 
and DESPOT~\citep{despot} similarly reduces the lookahead search tree by adaptive pruning. 
However, these methods do not use any kind of rollout with a base policy.
Furthermore, POMCP and DESPOT do not address  multiagent issues. 

On the other hand, various multiagent reinforcement learning and policy gradient methods~\citep{A3C,PPO,DQN_Rainbow} have been proposed. 
Among them, ~\citep{capitan2013decentralized} deals with multiagent cooperative planning under uncertainty in POMDP using decentralized belief sharing and policy auction, done after each agent executes a value iteration. 
The papers ~\citep{lowe2017multi, foerster2018counterfactual} consider an actor-critic policy gradient approach that scales well with multiple agents. 
However, the per-agent policy networks use only the local observations and do not leverage any extra information when the agents fully or partially communicate between themselves about their controls, observations, etc. 
By contrast, our methodology uses extra information whenever available. In Section~\ref{sec:resultsSOTA} we compare our performance to several of these state-of-the-art methods.

The paper ~\citep{Bhattacharya2020RL_PAPI} proposes rollout and PI methods that can address POMDP, but does not deal with multiagent problems, and has difficulty dealing with a large control space. The rollout cost improvement property given in  ~\citep{Bhattacharya2020RL_PAPI} also holds for the multiagent version of this paper. The proof was given in ~\citep{bertsekas2019multiagent} and an associated performance bound was given in the research monograph~\citep{Ber20}.

According to~\citep{grandChallengesWood}, robotic exploration is one of the biggest challenges in the field. The methods discussed in this paper are well suited for related multiagent contexts such as search and rescue applications~\citep{SARuavs, Cassandra2003,decisionMakingAutonomousVeh, controlRusUnderwater, serviceRobotGaurav,decisionMakingRus}.  Many researchers in robotics have proposed various methods for addressing robotic exploration and large state spaces ~\citep{ AIHowDeepRL, RLWolfram, deepLearningAIAbeel, AIDeliberation}, also in a POMDP setting~\citep{POMDPAmatoDecentralized, POMDPSycara}. The algorithms developed here are compatible with various coordination issues~\citep{gilMultiRobot,gilISRR2019, gilMultiRobotCov, controlRusUnderwater, POMDPAmatoDecentralized}.

\SG{This paper is most closely related to the multiagent rollout methods developed in~\citep{bertsekas2019multiagent} and autonomous repair problem in~\citep{Bhattacharya2020RL_PAPI}. Our work differs in various important ways from this prior work: 1) We treat the case of partially observable state, leading to an explosion in the size of the state space, not addressed in~\citep{bertsekas2019multiagent}, 2) we develop a multiagent decision-making framework in this partially observable context, leading to an explosion in the size of the action space, not treated in~\citep{Bhattacharya2020RL_PAPI}, 3) we consider more general environments described over arbitrary graph topologies (in contrast with the linear or strict grid topologies in \citep{Bhattacharya2020RL_PAPI}) and treat the significantly more challenging and realistic non-terminating case where previously fixed locations can fall into disrepair (not treated in~\citep{Bhattacharya2020RL_PAPI}), and finally in the appendix, 4) we consider cases where agents may not communicate their states and controls to one another explicitly.}

\vspace{-10pt}
\section{Belief Space Problem Formulation for Multiagent POMDP}
\vspace{-5pt}

We introduce the classical  belief space formulation of POMDP. We assume that there are $n$ states denoted by $i=1,\ldots,n$, and that the control $u$ consists of $m$ components, $u=(u_1,u_2,\dots,u_m)$. Each of the components corresponds to a separate agent. Given a starting state $i$, and control vector $u$, there is a known transition probability to reach the next state $j$, which is  denoted by $p_{ij}(u)$. Each component of the control $u_\ell, \ell\in\{1,2,\dots, m\}$, must belong to a finite  set $U_\ell$, so that the control space is the Cartesian product $U_1 \times U_2 \times \cdots \times U_m$.
The cost at each stage is denoted by $g(i,u,j)$ and is discounted with a factor $\alpha \in (0,1)$. The total cost is the sum of the $\a$-discounted expected costs incurred over an infinite horizon. 

\begin{wrapfigure}[13]{R}{0.4\textwidth}
	\vspace{-30pt}
	\begin{center}
		\includegraphics[width=\linewidth]{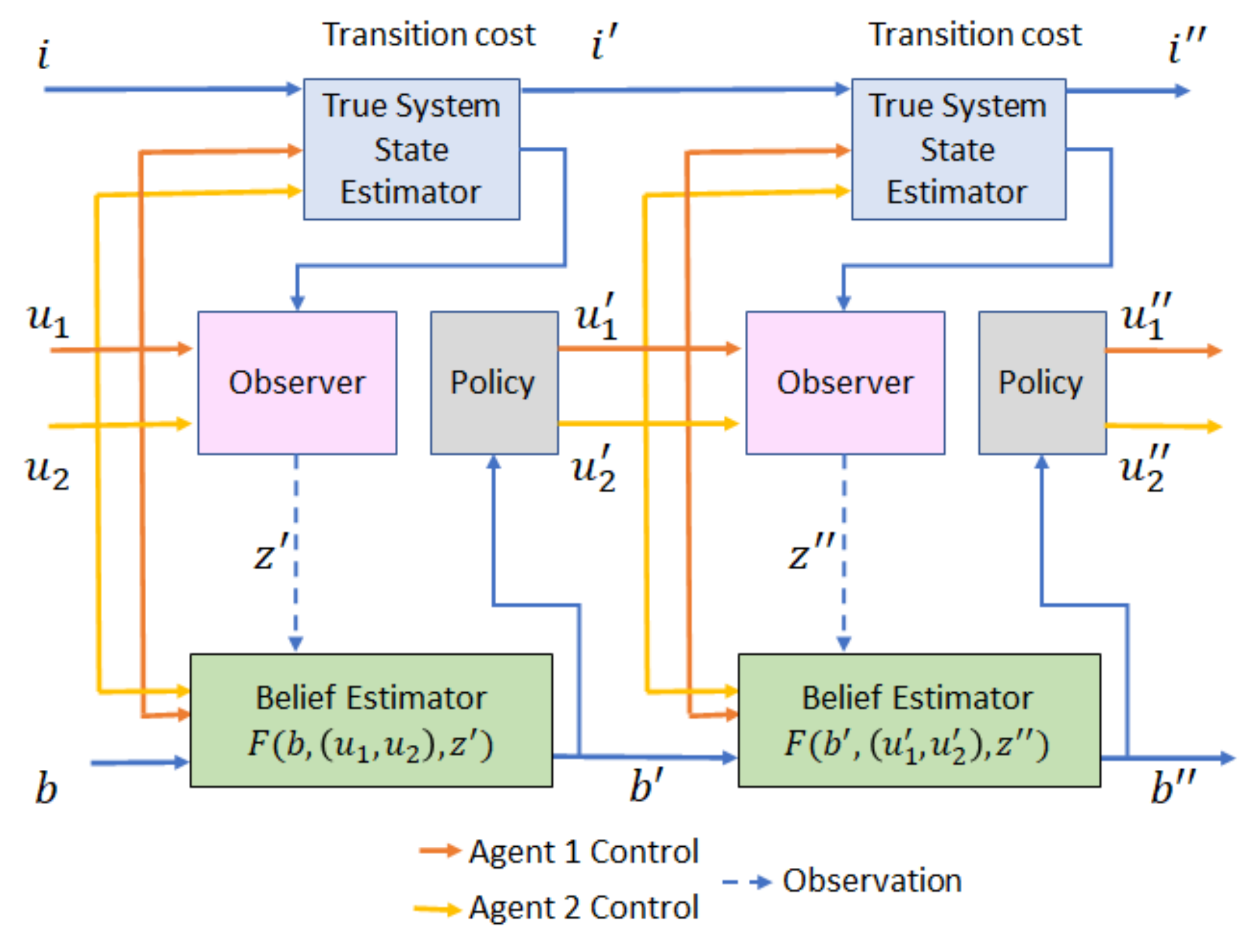}
	\end{center}
	\vspace{-15pt}
	\caption{\small Composite system simulator for POMDP for a given policy. The starting state $i$ of a trajectory is generated randomly using the belief state $b$ adopted from \citep{Bhattacharya2020RL_PAPI}.}
	\label{belief_simulator}
\end{wrapfigure}

We assume that a transition from state $i$ to the next state $j$ under control $u$, will generate an observation $z$ with a probability $p(z\mid j,u)$, where $z$ belongs to a known finite set $Z$.
However, we assume here that the agents share observations, so that all computations are done with full knowledge of the entire history of the observation vectors.  Our goal is to determine the control component for each agent at every stage as a function of the current belief state, which minimizes the discounted expected total cost,  starting from any initial belief state. 

We use the belief space transformation of a POMDP to a problem of perfect state information, which is the same as the one of~\citep{Bhattacharya2020RL_PAPI}. In particular, the belief state is the conditional probability vector $b=\big(b(1),\ldots,b(n)\big)$, where $b(i)$ is the conditional probability that the state is $i$, given the control-observation history up to the current time. 
The belief state  can be sequentially updated using a belief estimator $F(b,u,z)$, from a given belief state $b$, control $u$, and observation $z$ (see Fig.~\ref{belief_simulator}).

The optimal cost function $J^*(b)$ is the unique solution of the Bellman equation 
$$J^*(b) = \min_{u\in U}\left[\hat g\big(b,u\big)+\alpha \sum_{z\in Z}\hat p\big(z\,|\, b,u\big) J^*\big(F(b,u,z)\big)\right]\vspace{-0.1in}$$

Here $F$ is the belief state estimator, and 
$\hat{g}(b,u)$ and $\hat p(z\,|\, b,u)$ are defined by
\vspace{-1pt}
$$\hat g(b,u)=\sum_{i=1}^n b(i)\sum_{j=1}^n p_{ij}(u)g(i,u,j),\;\;\;\; \hat p(z\,|\, b,u)=\sum_{i=1}^nb(i)\sum_{j=1}^np_{ij}(u)p(z\mid j,u).$$
Our suboptimal solution approach is based on approximation in value space, implemented through the use of rollout. In particular, we replace $J^*$ in the Bellman equation with an approximation $\tilde J$. The corresponding suboptimal policy $\tilde{\mu}$ is obtained by the one-step lookahead minimization
\vspace{-5pt}
\begin{equation}
\label{eq:rollout}
\tilde \mu(b)\in\arg\min_{u\in U}\left[\hat g(b,u)+\alpha \sum_{z\in Z}\hat p(z\,|\, b,u) \tilde J\big(F(b,u,z)\big)\right].
\vspace{-0.08in}
\end{equation}
A more general version involves multistep lookahead minimization. In the pure form of rollout we use as $\tilde{J}$ the cost function of some policy, referred to as the base policy. In the next section, we define a rollout algorithm, which uses a simplified agent-by-agent lookahead minimization, and approximations $\tl J$ that involve a base policy with trajectory truncation and terminal cost approximation.

\vspace{-10pt}
	\section{Multiagent Truncated Rollout with Cost Function Approximation}
\begin{wrapfigure}[17]{R}{0.4\textwidth}
	 	\vspace{-40pt}
	 	\begin{center}
	 		\includegraphics[width=0.9\linewidth, height=6cm]{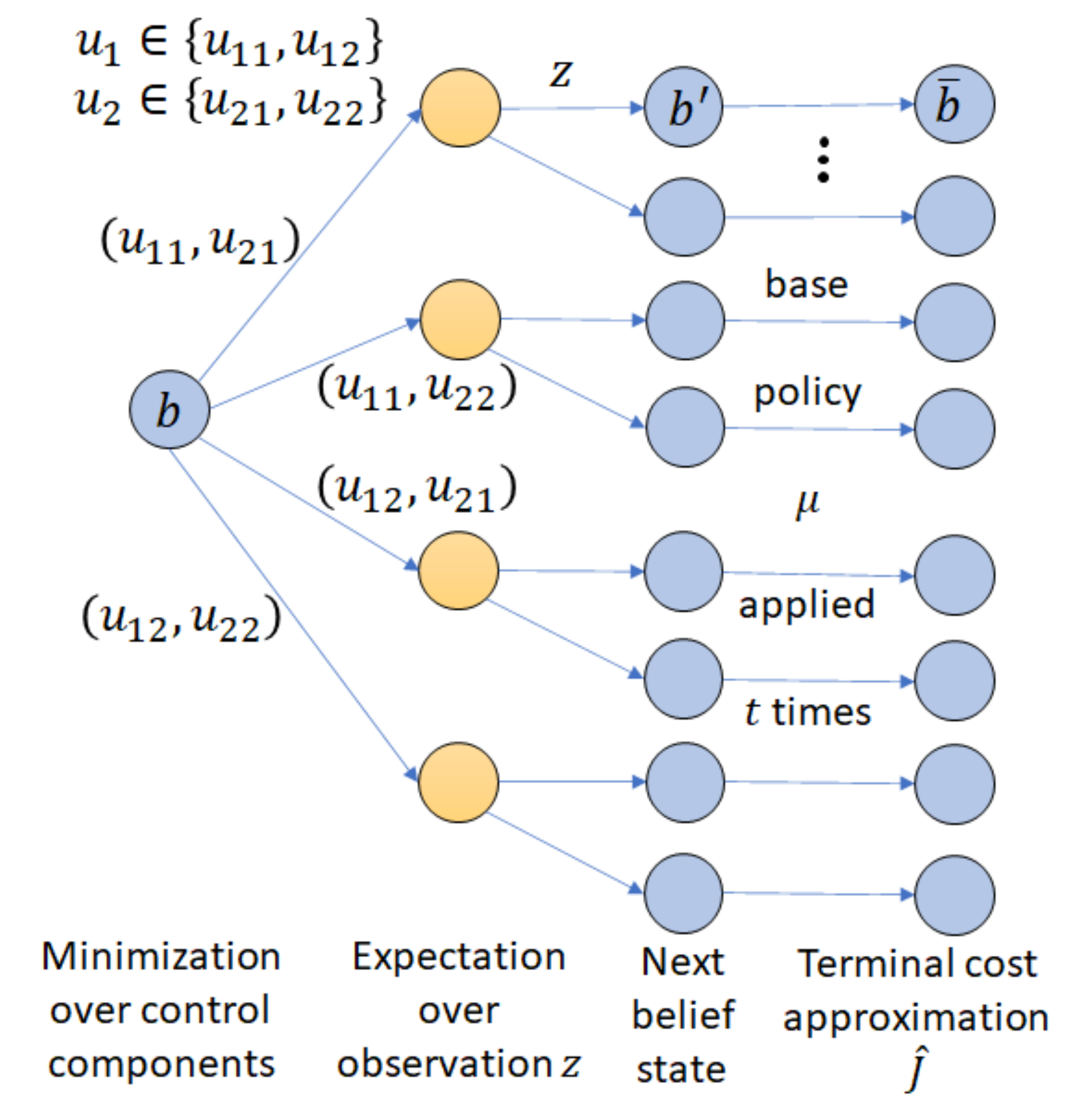}
	 	\end{center}
	 	\vspace{-0.15in}
	 	\caption{\small{Standard truncated rollout algorithm for $2$ agents: one-step lookahead followed by $t$-step application of the base policy $\mu$, and cost approximation $\hat{J}$.}}
	 	\label{fig:AATR}
	 	\vspace{-0.2in}
	 	\end{wrapfigure}

	\vspace{-10pt}
	 In the pure form of rollout with $l$-step lookahead, to find the rollout control at the current belief state $b$, we form an $l$-step lookahead tree using the transition and observation probabilities. Starting from each leaf node $b'$ of the tree, we use the  cost of the base policy $\mu$ as the cost approximation in  Eq. (\ref{eq:rollout}) [$\tilde{J}(b') = J_{\mu}(b')$]. In the truncated rollout version, $\tilde{J}(b')$ is the discounted cost of applying a base policy $\mu$ for a given number of $t$ stages, starting from the leaf node $b'$ and adding a terminal cost function approximation $\hat{J}(\bar{b})$, where $\bar{b}$ is the belief state obtained at the end of the $t$ steps of application of the base policy starting from $b'$. In other words, we truncate the system trajectory at the belief state $\bar{b}$ after $t$ stages, and we approximate the cost of the remainder of the trajectory with $\hat{J}(\bar{b})$; see Fig.~\ref{fig:AATR}. 
	 
	 The truncated rollout algorithm just described involves a few parameters: the lookahead length $l$, the length $t$ of the simulated trajectory before truncation, the choice of base policy $\mu$, and the terminal cost function approximation $\hat{J}$. The parameters $l$ and $t$ are usually chosen based on a tradeoff between implementation complexity and obtained performance. The base policy can be a greedy policy, and the terminal cost function approximation $\hat{J}(\bar{b})$ can be an estimate of the cost function of the base policy or an estimated steady-state cost from the belief state $\bar{b}$, or it may be simply set to 0.  The paper ~\citep{Bhattacharya2020RL_PAPI} (Prop.\ 1) provides theoretical performance bounds on the cost improvement of the truncated rollout algorithm. These bounds indicate among others that increasing the lookahead length $l$ improves the rollout performance bound. Moreover, they state that the performance of the rollout policy $\tilde\mu$ improves over the base policy $\mu$ as the terminal cost function approximation $\hat{J}$ gets closer to the base policy cost $J_{\mu}$. 
    \vspace{-10pt}
    \subsection{Standard rollout (\acrshort{sr})}
	\vspace{-5pt}
	In the standard form of rollout, at the current belief state $b$, we construct an $l$-step lookahead tree where each branch represents a possible control vector $u=(u_1,\ldots,u_m)$, where $u_\ell \in U_\ell$, $\ell=1,\ldots,m$ (see~\citep{bertsekas2019multiagent}). The branch corresponding to control $u$ is associated with a Q-factor corresponding to $(b,u)$, which is the expression in brackets in Eq.\ (1). The standard rollout algorithm chooses the control that is associated with minimal Q-factor, cf.\ Eq.\ (1). 
	This rollout algorithm possesses the cost improvement property $J_{\tilde{\mu}}(b) \le J_{\mu}(b)$, where $\mu$ is the base policy and $\tilde{\mu}$ is the rollout policy.  
	The difficulty with this formulation is that at each stage the overall rollout algorithm computation is of order $O(C^m)$ for each stage, where $C=\max\{|U_1|,|U_2|,\dots,|U_m|\}$ is the maximum cardinality of the control component constraint sets. To alleviate this difficulty, we will introduce next a multiagent variant of rollout, where the lookahead minimization is performed one agent at a time, and the computation at each stage is reduced to $O(Cm)$. Fig.~\ref{fig:AATR} demonstrates standard rollout with two agents, each having two possible control components.
    \vspace{-10pt}
    \subsection{One-agent-at-a-time rollout (\acrshort{mr})}
	\vspace{-5pt}
	In order to reduce the algorithmic complexity of the standard rollout algorithm, the minimization over the control branches in the above formulation needs to be simplified. To achieve improved algorithmic complexity, we introduce an equivalent problem formulation where the control $u = (u_1, u_2, \dots, u_m)$ is broken down into its $m$ components. That is, given a belief state $b$, $m$ intermediate states are generated such that the agents choose their control components sequentially between the current belief state $b$ and the next belief state $b'$. Thus, the transition sequence from $b$ and $b'$ is $\big\{b, (b, u_1), (b, u_1, u_2), \dots, (b, u_1, u_2, \dots, u_{m-1}), b'\big\}$ assuming the agents choose their controls sequentially in a fixed order. The last transition from $(b, u_1, u_2, \dots, u_{m-1})$ to $b'$ involves the choice of the last component $u_m$ and includes the cost $\hat g(b,u)$ of choosing control $u=(u_1, u_2, \dots, u_{m})$ at the current belief state $b$ [cf.\ Eq.\ (1)]. Every other intermediate transition has $0$ cost.
	
\begin{wrapfigure}[11]{R}{0.7\textwidth}
		\vspace{-35pt}
		\begin{center}
		\includegraphics[width=\linewidth]{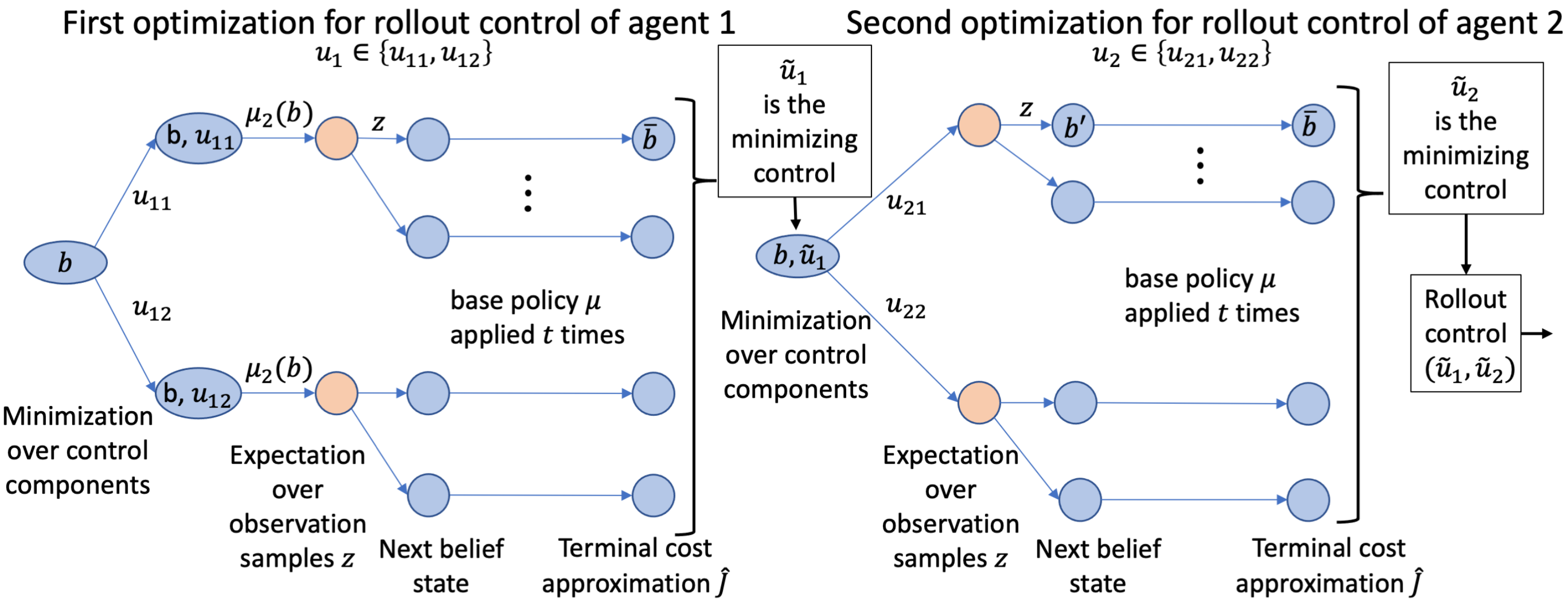}
		\end{center}
		\vspace{-0.15in}
		\caption{\small 1-at-a-time truncated rollout algorithm for $2$ agents using base policy $\mu$ with terminal cost approximation $\hat{J}$ on the reformulated state space $(b),(b,\tilde{u}_1)$.}
		\label{fig:OAATR}
	\end{wrapfigure}

	In the reformulated problem, at each stage, the rollout algorithm performs $m$ sequential optimizations over Q-factors that involve a single control component.
In particular, when optimizing over the Q-factors of component $u_\ell$, we set $u_1,\ldots,u_{\ell-1}$ to the optimized values calculated earlier by the rollout algorithm, and we set $u_{\ell+1},\ldots,u_{m}$ to the values dictated by the base policy at the current belief state.
	The per-stage complexity of this rollout algorithm is $O(Cm)$, which can be a dramatic improvement over the exponential computational complexity of the (all-at-once) standard rollout algorithm. Our computational experiments are consistent with the results of other studies, namely that this computational economy is often obtained with minimal loss of performance. The algorithm is illustrated in Fig.~\ref{fig:OAATR}.
	
	In the paper~\citep{bertsekas2019multiagent} and the research monograph~\citep{Ber20}, the one-agent-at-a-time rollout method was shown to maintain the cost improvement property of standard rollout. For the case of one-step lookahead, it was also shown that the performance bounds for the standard and the one-agent-at-a-time truncated rollout algorithms are identical (see Prop.\ 5.2.7 of ~\citep{Ber20}). 
\vspace{-10pt}
\subsection{Order-optimized rollout (\acrshort{oomr})}
	\vspace{-5pt}
	The preceding rollout algorithm assumes a fixed a priori chosen order in which the agent control components are optimized. However, the algorithm also works with any agent order, and in fact, it also works if the order is changed at each stage. This motivates algorithmic variants where the agent order is approximately optimized at each stage. 
	An effective and relatively inexpensive way to do this is to first optimize over all single agent Q-factors, by solving the $m$ minimization problems that correspond to each of the agents $\ell=1,\ldots,m$ being first in the one-agent-at-a-time rollout order. If $\ell_1$ is the agent that produces the minimal Q-factor, we fix $\ell_1$ to be the first agent in the one-agent-at-a-time rollout order. Then we optimize over all single agent Q-factors, by solving the $m-1$ Q-factor minimization problems that correspond to each of the agents $\ell\ne \ell_1$ being second in the one-agent-at-a-time rollout order. If $\ell_2$ is the agent that produces the minimal Q-factor, we fix $\ell_2$ to be the second agent in the one-agent-at-a-time rollout order, and continue in this manner. In the end, after 
${m(m+1)/2}$ minimizations, we obtain an agent order $\ell_1,\ldots,\ell_m$ that produces a potentially reduced Q-factor value, as well as the corresponding rollout control component selections. 
Based on our experimental results, agent order optimization produces modest, but significant and consistent performance improvement over the case of a fixed agent order.

\vspace{-15pt}
	\section{Multiagent Approximate Policy Iteration}
	\vspace{-10pt}
	\begin{wrapfigure}[11]{r}{0.45\textwidth}
	\vspace{-30pt}
	\begin{center}
		\includegraphics[width=0.9\linewidth]{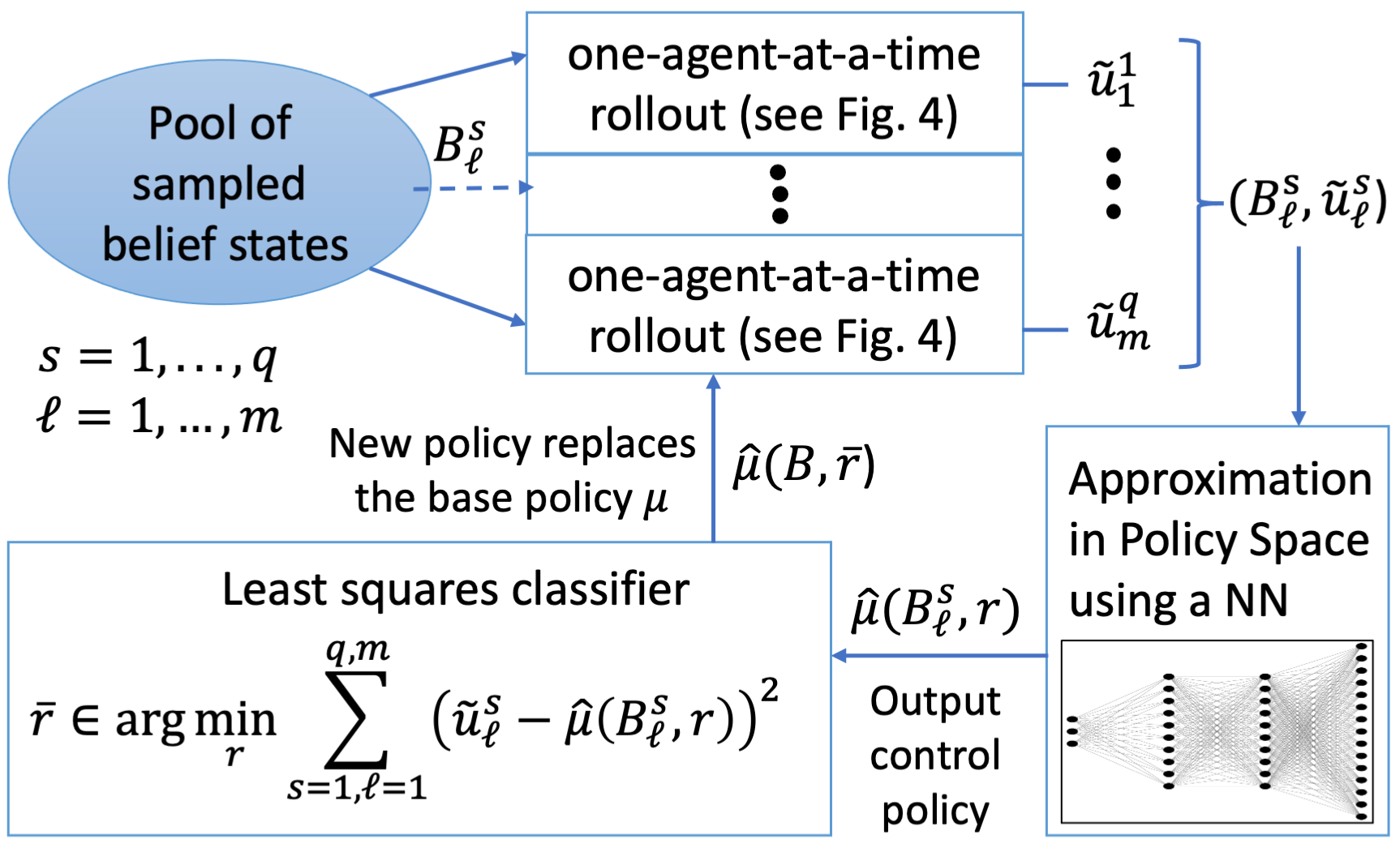}
	\end{center}
	\vspace{-0.15in}
	\caption{\small Approximate PI algorithm based on multiagent rollout and approximation in policy space. Adapted from~\citep{Bhattacharya2020RL_PAPI}.\label{fig:NN}}
\end{wrapfigure}
	We will now discuss the approximate PI method as an extension to the rollout algorithm. 
	The truncated rollout policy can be considered as the base policy in PI. 
	The policy evaluation is done in an online fashion by $l$-step lookahead minimization over the simulated trajectories (using base policy $\mu$, $t$ times followed by the terminal cost approximation $\hat{J}$), at each stage. 
	The subsequent iterations can be expedited by replacing the online evaluation of the rollout policy with an approximation architecture (namely a neural network) as shown in Fig.~\ref{fig:NN}. See~\citep{Ber19}, Sections 2.1.5 and 5.7.2. Here, the newly trained approximation architecture for the rollout policy serves as the subsequent base policy for the next iteration. Using these multiagent truncated rollout schemes as a basis, we now describe corresponding approximate PI algorithms.

	\vspace{-10pt}
	\paragraph{Approximate PI with truncated rollout:}This algorithm uses standard rollout to generate the belief state-rollout control pairs to train the policy network in each iteration. 
		We define a parametric policy approximation $\hat{\mu}(b, r)$, that produces a control given a belief state $b$, where $r$ is the parameter of the approximation architecture. For example, a neural network can be trained with a large set consisting of $q$ belief state - control pairs $(b^s,u^s)$, $s =1,\dots,q$, in a supervised learning fashion, where $r$ may include the weights of each layer. We can estimate the rollout control $u^s$ from a belief state $b^s$ and add it to the training set. 
		The training process solves the classification problem using the training set and generates a neural network-based approximation $\hat{\mu}(b,\bar{r})$ for the rollout policy, which in turn is used as the base policy for the next iteration. This idea was proposed in the context of PI in the paper~\citep{LaP03}, and is also described in the book~\citep{Ber19}, Section 3.5. This is expensive in terms of computation especially with a large number of agents, using $C^m$ Q-factors, and thus is not applied directly here.  Instead, we extend upon this idea for the one-agent-at-a-time case discussed next. 
		
		\vspace{-10pt}
		\paragraph{Approximate PI with truncated one-agent-at-a-time rollout (\acrshort{mr_api}):}This algorithm uses the one-agent-at-a-time rollout scheme to train the parametric architecture for policy space approximation. Given a belief state $b^s$, the one-agent-at-a-time rollout algorithm produces one agent's control component $\tilde{u}^s_{\ell}$, $\ell\in{1,2,\dots,m}$ at a time. Each component of the rollout policy, starting from $\ell=\{1,2,\dots,m\}$ is given by the following equation at each stage:
		 \vspace{-0.1in}
		\begin{equation*}
		\tilde u^s_\ell \in
		\arg\min_{u\in U_\ell}\hat g(b^s,u'^s) + \alpha \sum_{z\in Z}\hat p(z\,|\, b^s,u'^s) J_{\mu}\big(F(b^s,u'^s,z)\big) 
		\vspace{-0.1in}
		\end{equation*}
		where $u'^s =(\tilde{u}_{1:\ell-1}^s,u,u_{\ell+1:m}^s)$, and $u^s_\ell$ denotes the base policy's control component for the $\ell^{th}$ agent. 
		At the end of $m$ such optimizations we construct the entire rollout control $\tilde{u}^s$. All pairs of $(B^s_\ell,\tilde{u}^s_\ell)$, where $B^s_\ell = (b^s,\ell,\tilde{u}_{1:\ell-1}^s,u_{\ell+1:m}^s)$ are used to train the approximation architecture and obtain the policy network $\hat{\mu}(B,\bar{r})$. The policy network is trained with $qm$ samples in total. 
		To construct the entire control vector from the approximate architecture in an iteration, we need to invoke the policy network $m$ times. For example, when estimating the first component  ($\tilde{u}_1$) of the rollout control $\tilde{u}$ for a belief state $b$, we need to construct tuple $B_1=(b,1,u_{2:m})$, where $u=(u_1,\dots,u_m)$ denotes the base policy's control components at $b$.  Invoking the policy network $\hat{\mu}(B_1,\bar{r})$, will produce the first rollout control component $\tilde{u}_1$. The second component $\tilde{u}_2$ of the rollout control is similarly estimated by first constructing the tuple $B_2=(b,2,\tilde{u}_1,u_{3:m})$ and invoking the policy network $\hat\mu(B_2,\bar{r})$. Repeating this process will produce the entire rollout control (see Fig.~\ref{fig:NN}).
\vspace{-10pt}
		\paragraph{Approximate PI with truncated order-optimized rollout:}Similar to the previous variant, one can use the order-optimized rollout scheme to generate the belief state - rollout control pairs and train the approximate policy network. Apart from that, this variant of approximate PI will exactly follow the previous approximate PI method. Note that one-agent-at-a-time rollout (as opposed to standard rollout)
		has been used in all of our approximate PI experiments.

%
%

\vspace{-10pt}
\section{Comprehensive Simulation Studies and Comparative Results}
\label{sec:result}
\vspace{-10pt}
\begin{wrapfigure}[13]{r}{0.6\textwidth}
	\vspace{-22pt}
	\begin{center}
		\includegraphics[width=\linewidth]
		{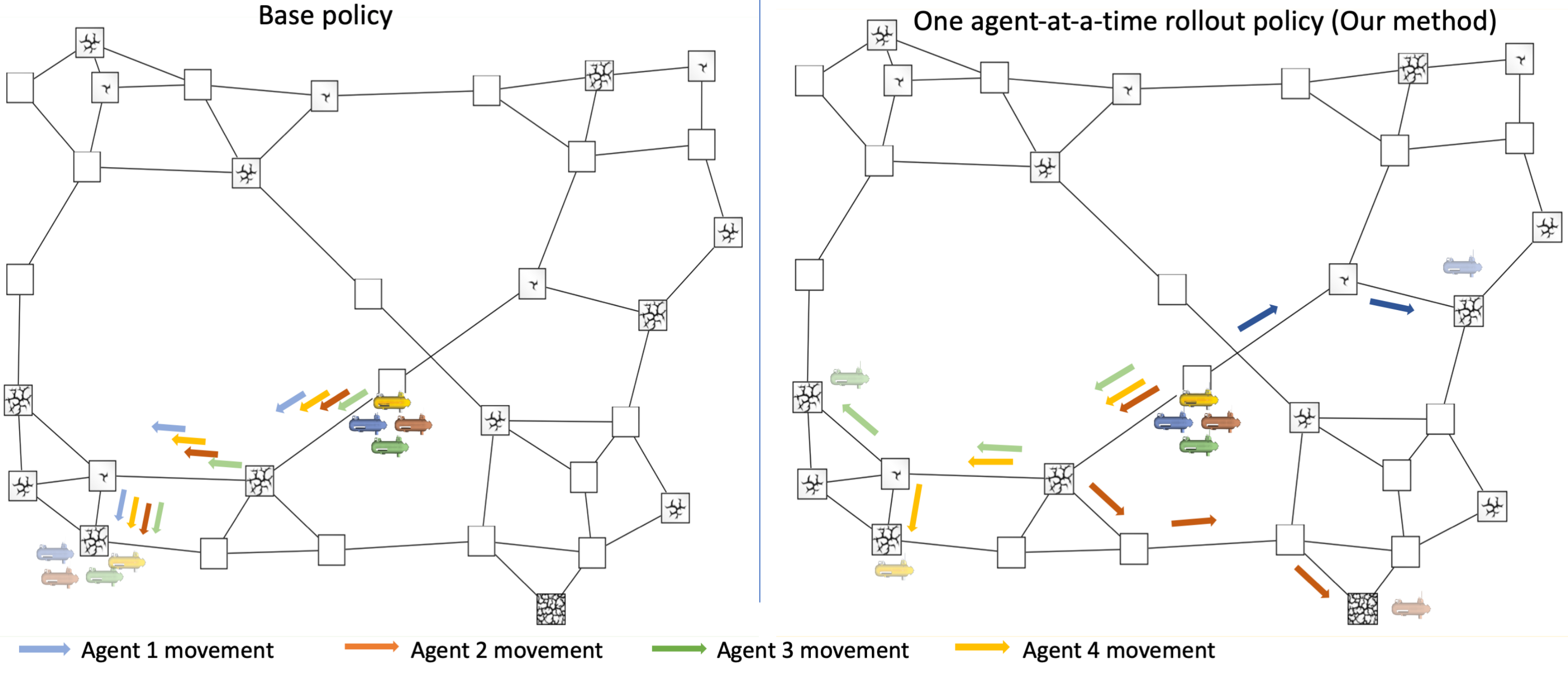}
	\end{center}
	\vspace{-12pt}
	\caption{\small Trajectories generated by a base policy (left) vs our 1-at-a-time rollout policy (right) on a 2D repair network. 
	Note the coordinated splitting behavior of the 1-at-a-time rollout policy in contrast to the base policy.\label{fig:traj1}}
    \end{wrapfigure}
\begin{wrapfigure}[9]{R}{0.38\textwidth}
		\centering
		\vspace{-25pt}
		\includegraphics[width=0.8\linewidth]{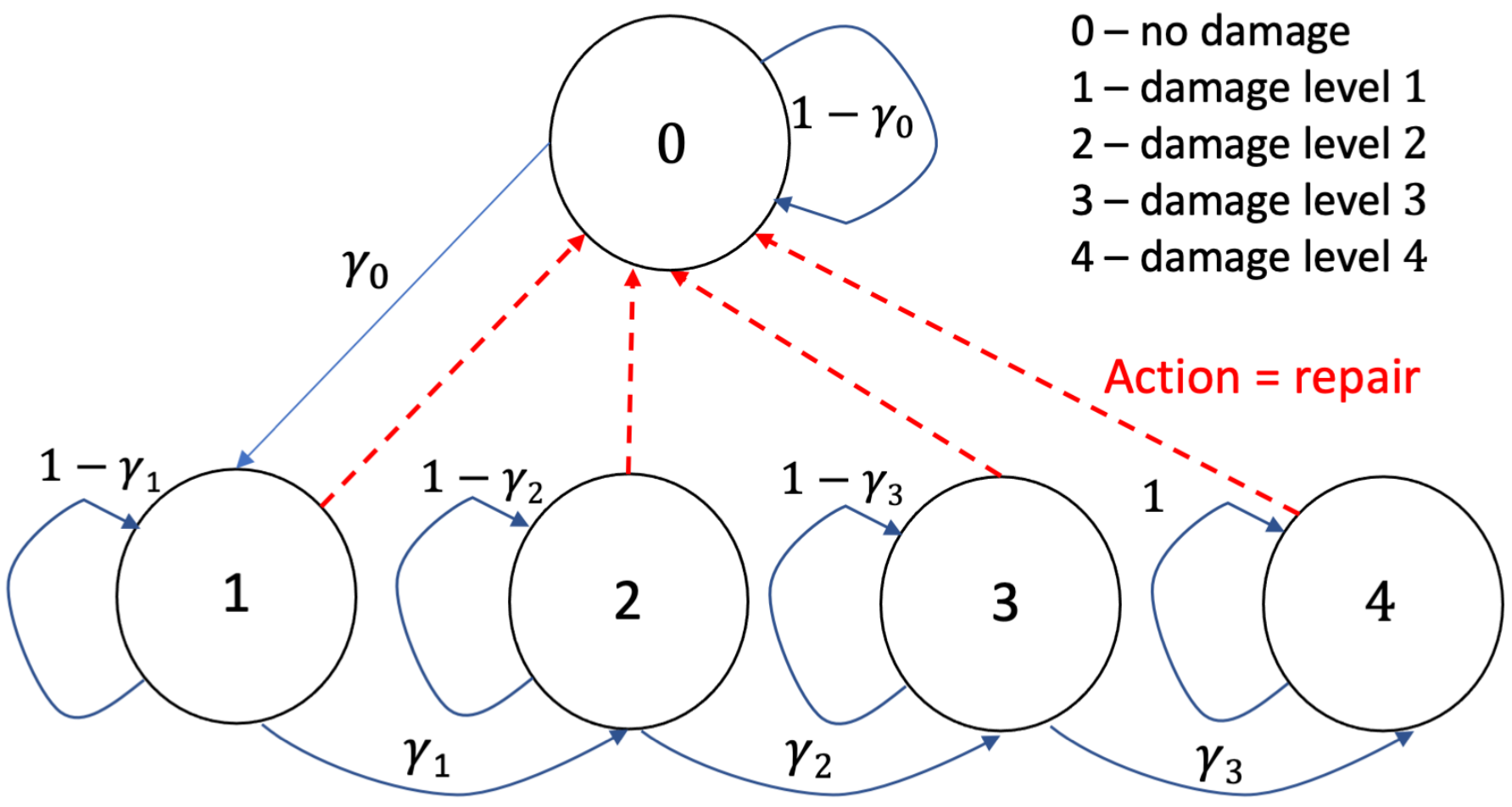}
		
		\vspace{-8pt}
		\caption{\small Markov chain for the damage level of each network location. Generalization from~\citep{Bhattacharya2020RL_PAPI} for a 2D 
		topology where fixed locations can fall into disrepair.
		}
		\label{fig:markov}
	\end{wrapfigure}
\begin{wrapfigure}[11]{R}{0.60\textwidth}
		\centering
		\vspace{-17pt}
	\begin{minipage}{0.3\textwidth}
		\centering
		\captionsetup{width=\linewidth}
		\vspace{-0pt}
		\includegraphics[width=\linewidth]{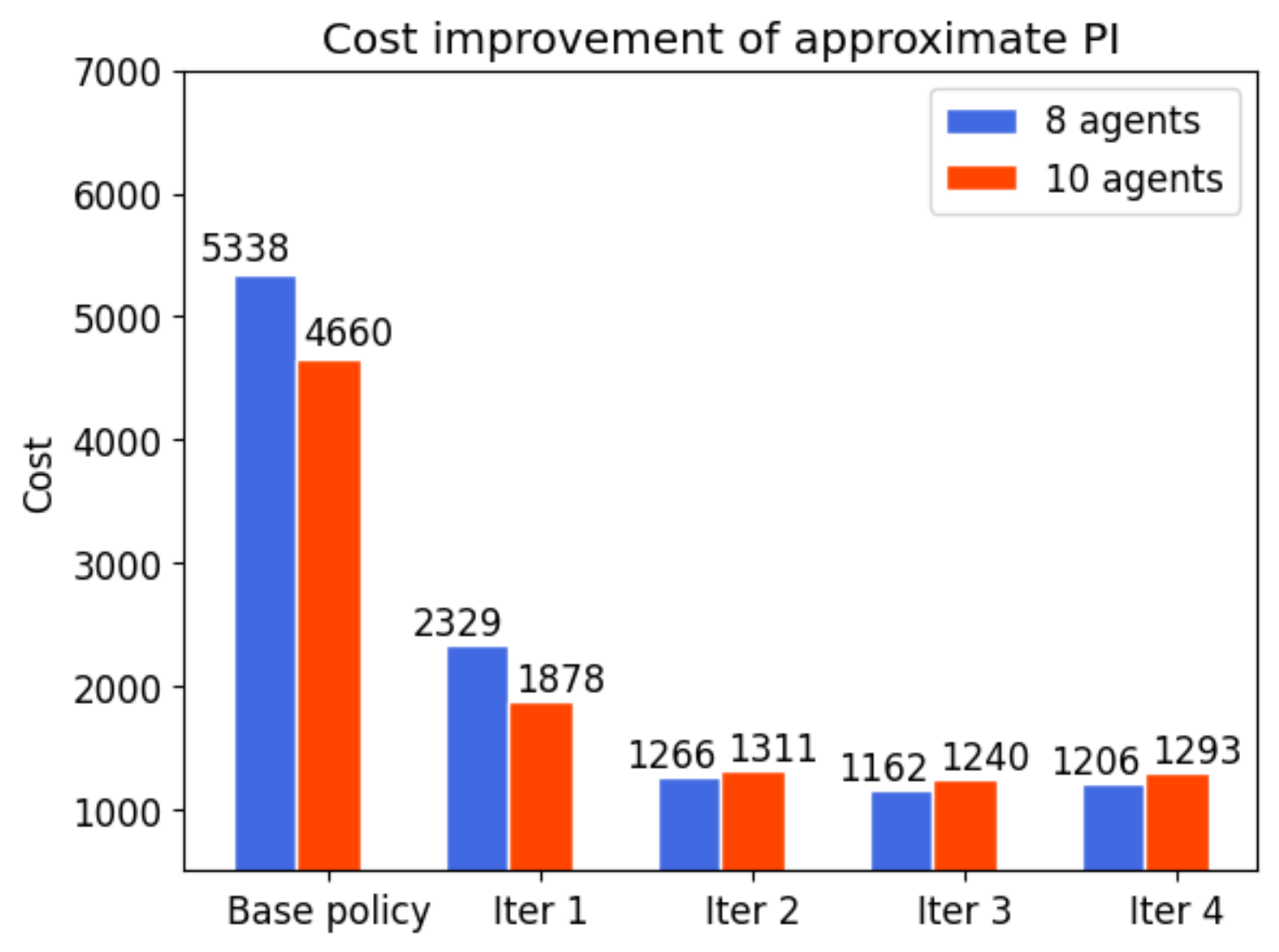}
		\vspace{-18pt}
		\caption{\small{Cost of base policy and approximate PI with 1-at-a-time truncated rollout.}
		}
		\label{fig:api_results_10}
	\end{minipage}
	\begin{minipage}{.29\textwidth}
			\centering
		\captionsetup{width=\linewidth}
			\hspace{-0.12in}
			\includegraphics[width=\linewidth]{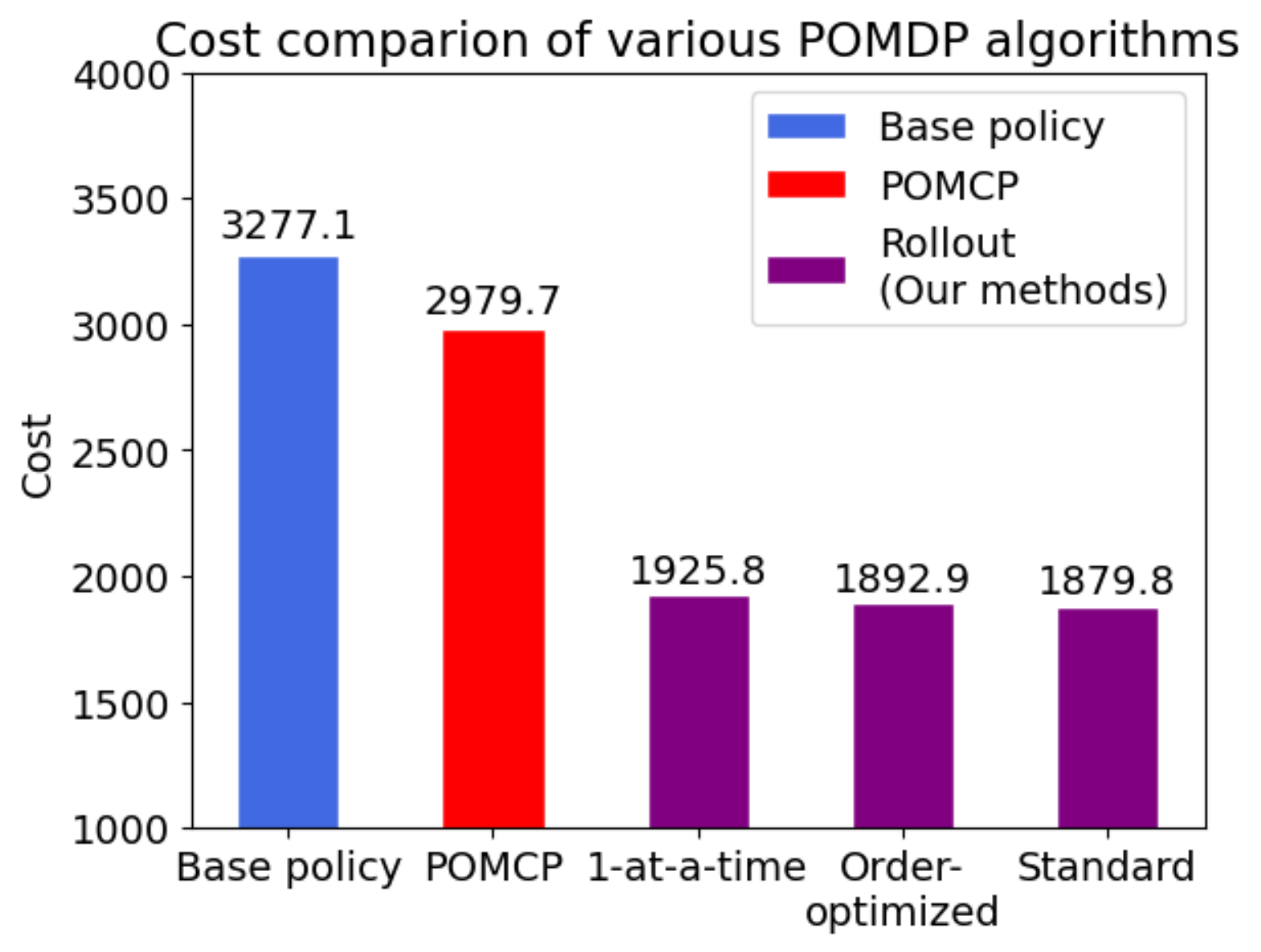}
		\vspace{-7pt}	\caption{\small{Cost comparison of POMCP, base policy, and rollout algorithms with 4 agents.}
			}
		\label{fig:rollout_results}
		\end{minipage}
		\vspace{-15pt}
\end{wrapfigure}

	In this section, we provide computational results and a comparative study with existing POMDP methods. Our computational results demonstrate that: 1) our one-agent-at-a-time rollout and order-optimized rollout result in substantial computational savings with comparable performance vs. the standard rollout, 2) our one-agent-at-a-time approximate PI method improves the policy over several iterations, 3) our methodologies applied to a complex multi-robot repair problem significantly outperform existing methods, and work well where other methods fail to scale up (e.g. with 10 agents), 4) use of a rollout policy naturally gives rise to coordination among agents even when this coordination does not result from a straightforward application of its base policy (see Fig.~\ref{fig:traj1}), and \SG{5) one-agent-at-a-time rollout with imperfect communication (see Appendix sections~\ref{sec:communication_less},\ref{sec:communication}) demonstrates qualitatively similar performance to the perfect communication case under certain conditions}.
	\vspace{-10pt}
	\paragraph{Multi-robot repair problem:} 
	We are interested in solving a challenging multi-robot repair problem on a partially observable network with several damaged nodes. Our objective is to learn a coordinated policy for the agents resulting in repair with minimum cost. Here, damaged node locations can be a proxy for many applications from pipeline damage locations, to forest fire threats, to damaged equipment sites in a power grid. The network is represented as an undirected graph with vertex set $V$ denoting locations and each location $v \in V$ has one of $\nu$ damage levels $(0,1,\ldots,\nu-1)$ that evolve over time according to a known Markov Decision Process (MDP) with $\nu$ states as shown in Fig.~\ref{fig:markov}. A fixed location (with damage level $0$) can become damaged over time as shown in Fig.~\ref{fig:markov} by a non-zero transition probability from $0$ to $1$. This problem is a generalized extension of the pipeline problem discussed in the paper~\citep{Bhattacharya2020RL_PAPI} where the 2D topology and possible decay of a repaired location make this problem a significantly more difficult infinite horizon problem than that described in~\citep{Bhattacharya2020RL_PAPI}.
	We assume perfect communication and perfect observations on the damage levels of the agents' current locations. The damage distribution for each location $v$ can be represented as $d^v=(d_0^v,\ldots,d_{\nu-1}^v)$ consisting of the conditional probabilities 
	of the damage level given the prior initial belief and a control-observation history for all agents.
	The shared belief consists of the locations of all agents and damage distributions of all locations in the graph.
	At each time step, once an agent at location $v$ has made an observation with respect to its current location, it can either choose to stay in $v$ and fix the location or move to one of its neighboring locations. 
	The agents incur a cost per unit time $\tau$ whenever there is nonzero damage in the network, denoted $C_\tau = \sum_{ v \in V} {d^v} \cdot {c}$, where ${c}$ is a cost vector that maps a damage level to a cost 
	(we assumed $c=[0,0.1,1,10,100]$ in our experiments). The policy needs to minimize the discounted sum of costs over an infinite horizon ($\sum_\tau \alpha^\tau C_\tau$). This is a POMDP with $|V|^m \nu^{|V|}$ states and a variable control space depending on the location.
	As a terminal cost approximation we have used a steady-state value which is the discounted cost sum over an infinite horizon assuming that no further control is applied, $\hat{J} = (1/(1-\alpha))\sum_{v\in V} {d^v}{c}$. This type of approach works well when the lookahead tree has a high branching factor, or when simulating the trajectories is fairly expensive.
	\vspace{-10pt}
	\paragraph{Simulation setup:} 
	We implement the multiagent rollout methods 
	on a graph topology (shown in Fig.~\ref{fig:graph}) with $32$ nodes and $4,8,10$ agents (state space size $10^{28}, 10^{34}, 10^{37}$, and control space size $625,10^{5.6},10^7$, respectively). We use a discount factor $\alpha=0.95$.
	A variant of the problem where a repaired location remains fixed, is significantly easier for the agents to solve and this has been used for comparative studies with other existing methods ($\alpha=0.99$ for this variant). Table \ref{table:rollout2_results}, 
	Fig.~\ref{fig:api_results_10}, and Fig.~\ref{fig:rollout_results} use the values $\gamma_0=0.01$ ($0$ for 4 agents), $\gamma_1=0.02, \gamma_2=0.03, \gamma_3=0.05, \gamma_4=0.1$. The base policy for each agent \SG{is chosen to be a relatively simple ``greedy policy'' that does not require any problem-specific tailoring, whereby it chooses to fix the current location (if damaged) and otherwise takes one step towards the nearest damaged location.} We use Dijkstra's shortest path algorithm to determine the nearest damaged location and the next hop from each location. We run the rollout and API algorithms on MPI enabled ASU Agave cluster with Intel Xeon E5-2680 v4 CPUs (56, 196 cores respectively). All costs reported in the result section are aggregated over $1000$ random initial states.
    \vspace{-10pt}
	\paragraph{Performance of multiagent rollout:}
	Table \ref{table:rollout2_results} shows the cost comparison between base policy and one-agent-at-a-time rollout policy. 
   The results show that the one-agent-at-a-time rollout performed significantly better than its base policy. 
    This behavior is consistent with the rollout cost improvement property. Fig.~\ref{fig:traj1} demonstrates representative agent trajectories resulting from our rollout policy which reveals coordination behavior learned by the agents. Specifically, agents learn to split their efforts to tackle the repair problem most efficiently (Fig.~\ref{fig:traj1} right side), leading to improved performance.  This is in contrast to the base policy where agents duplicate repair efforts by moving through the graph in concert (Fig.~\ref{fig:traj1} left side). An alternative initial scenario involves initiating two agents in two different sections of the graph where one was more severely damaged than the other. In this case, a base policy keeps the agents in their corresponding initial sections leading to longer repair times. When using one-agent-at-a-time rollout in contrast, the agent starting in a mildly damaged section moves to the most damaged section to assist other agents. 
	\begin{wraptable}{R}{35mm}
	\vspace{-28pt}
    \centering
    \caption{\small Cost comparison of base policy, and 1-at-a-time rollout policy}
    \vspace{-5pt}
    \label{table:rollout2_results}
    
    \begin{tabular}{|p{5mm}|p{6mm}|p{11mm}
    |}
    \hline
    \scriptsize
    Agent&\scriptsize Base&\scriptsize{1-at-a-time}\\
        \hline
         8 &5347 & 992 \\ \hline
         10 & 4667 & 799 \\
         \hline
    \end{tabular}
    \vspace{-10pt}
\end{wraptable}
\vspace{-10pt}
\paragraph{Performance of multiagent approximate PI:} The neural network used for policy space approximation in this method has two hidden layers (with 256 and 64 ReLU units respectively) followed by a batch-norm layer. The output layer is a softmax layer which provides the probability distribution over the control components for an agent. The size of the output layer is $|V|+1$ (one control component is to fix the current location, and others represent the likelihood of traveling to each node, one likelihood value for each $v \in V$). We use RMSProp optimizer (learning rate = 0.001). We use a one-agent-at-a-time rollout
(with $l=1,t=10$) for policy improvement at each iteration. 

We use $500000$ training samples to train the policy network in each iteration. The training samples were generated by choosing a random set of belief states, followed by sampling from a memory buffer. Note that exploration issues are one of the main challenges in this context, and various solutions have been proposed to resolve this issue; see ~\citep{LaP03, DiL08}. To this effect, our memory buffers consist of states generated by taking a few steps from the initial state pool using one of the previous policies and a randomized policy; 
see~\citep{Ber19}, Ch.\ 5. Fig.~\ref{fig:api_results_10} shows
performance of neural network policies generated by approximate PI with $8,10$ agents respectively.
The results show that even with a large state and control space, approximate PI with one-agent-at-a-time rollout retains its cost improvement property over several iterations.

	\vspace{-10pt}
	\paragraph{Performance comparison to existing methods:}
	\label{sec:resultsSOTA}
	This section presents cost comparisons of a base policy, several 
	existing methods, standard rollout, our 
	one-agent-at-a-time rollout, and our order-optimized rollout. Note that here we cap the number of agents at 4 due to scalability issues (explosion in the number of Q-factor evaluations) for several of the methods including standard rollout and POMCP~\citep{SiV10}. In fact, due to scalability issues DESPOT~\citep{despot} was unsuccessful for the $4$-agent problem within a reasonable time limit ($1$ second per stage). For the same time limit, POMCP was able to deal with the $4$ agent case, but not a larger number of agents due to a combination of scalability problems involving computation time and memory requirements. 
	Fig.~\ref{fig:rollout_results} shows that standard rollout outperformed all other approaches and our one-agent-at-a-time rollout methods performed comparably well with dramatically less computation (this performance behavior was observed on a broad range of tests involving up to $4$ agents). 
	At the same time, the standard rollout method could not solve the problem with $8$ and $10$ agents due to the scalability issue.
	Furthermore, as expected, order-optimized rollout outperforms one-agent-at-a-time rollout. Notably, all the variations of our rollout algorithms perform significantly better than the base policy. 
	
	We compare our methods with two 
	existing 
	learning methods, POMCP~\citep{SiV10} and MADDPG~\citep{lowe2017multi}. POMCP uses MCTS-based lookahead. For our implementation we use default parameters for POMCP (given in~\citep{despot}) and we modify the code to use a closed form of the belief update governed by the Markov chain in Fig.~\ref{fig:markov} (with $\gamma_0=0$). A single particle with weight=1 was used to represent the belief.
	Fig.~\ref{fig:rollout_results} shows the cost comparison of our methods with POMCP which improves the base policy but performs worse than our multiagent rollout methods.
	One of the reasons is that using a long and sparse lookahead tree results in poor Q-factor estimation in problems with a long planning horizon. In contrast, our rollout methods use shorter lookahead and more precise Q-factor estimation by simulations.
\SG{We used both the public source code provided by the MADDPG authors and the Berkeley Ray RLLib implementation of MADDPG, and conducted an extensive hyperparameter search to tune its parameters. However, it was consistently outperformed by the base policy for this multiagent repair problem} 
	and produced a cost $5520$ (compared with $3277$ for a base policy; see Fig.~\ref{fig:rollout_results}). 
	
	

\vspace{-10pt}
\section{Conclusion}
\label{sec:conclusion}
\vspace{-10pt}
In this paper we present various multiagent rollout methods and approximate PI for challenging large-scale POMDP problems. We experimentally verify the cost improvement property of one-agent-at-a-time rollout, similar to standard rollout, with dramatically less computation requirements. Similarly, we show that multiagent approximate PI improves the policy at each iteration in order to find the approximately optimal policy. Agents executing the resulting policy achieve a high degree of coordination with each other. \SG{We also present extensions of our multiagent rollout methods and report numerical performance results, for the imperfect communication case.}  Based on our experimentation, the methods discussed here work well for robotics problems particularly when a large team of multiple robots need to collaborate on a complex task over long horizons, a large state space with partial observations, and \SG{a large action space (induced by a large number of agents)}.
Future extensions to our POMDP algorithms include but are not limited to asynchronous communication of the belief state, imperfect knowledge of the agents’ locations, and partitioning of the belief state space to achieve distributed learning.
\vspace{-0.1in} \acknowledgments{\vspace{-0.1in} We thank the reviewers who provided useful comments, to colleagues who contributed to the ideas, and we gratefully acknowledge partial funding through the NSF CAREER grant number 1845225.}




\bibliography{example}  

\begin{thebibliography}{38}
\providecommand{\natexlab}[1]{#1}
\providecommand{\url}[1]{\texttt{#1}}
\expandafter\ifx\csname urlstyle\endcsname\relax
  \providecommand{\doi}[1]{doi: #1}\else
  \providecommand{\doi}{doi: \begingroup \urlstyle{rm}\Url}\fi

\bibitem[Bertsekas(2020{\natexlab{a}})]{bertsekas2019multiagent}
D.~P. Bertsekas.
\newblock Multiagent {Rollout Algorithms and Reinforcement Learning}.
\newblock \emph{arXiv preprint arXiv:1910.00120}, April 2020{\natexlab{a}}.

\bibitem[Bertsekas(2020{\natexlab{b}})]{Ber20}
D.~P. Bertsekas.
\newblock \emph{Rollout, Policy Iteration, and Distributed Reinforcement
  Learning}.
\newblock Athena Scientific, Belmont, MA, 2020{\natexlab{b}}.

\bibitem[Bertsekas(2019)]{Ber19}
D.~P. Bertsekas.
\newblock \emph{Reinforcement {Learning and Optimal Control}}.
\newblock Athena Scientific, Belmont, MA, 2019.

\bibitem[Silver and Veness(2010)]{SiV10}
D.~Silver and J.~Veness.
\newblock Monte-{Carlo Planning in Large POMDP}s.
\newblock In \emph{Proc. 23rd International Conf. on NeurIPS}, pages
  2164--2172, Red Hook, NY, USA, 2010.

\bibitem[Lowe et~al.(2017)Lowe, Wu, Tamar, Harb, Abbeel, and
  Mordatch]{lowe2017multi}
R.~Lowe, Y.~Wu, A.~Tamar, J.~Harb, P.~Abbeel, and I.~Mordatch.
\newblock Multi-{Agent Actor-Critic for Mixed Cooperative-Competitive
  Environments}.
\newblock \emph{Neural Information Processing Systems (NIPS)}, 2017.

\bibitem[Kaelbling et~al.(1998)Kaelbling, Littman, and Cassandra]{KLC98}
L.~P. Kaelbling, M.~L. Littman, and A.~R. Cassandra.
\newblock Planning and acting in partially observable stochastic domains.
\newblock \emph{Artificial intelligence}, 101:\penalty0 99--134, 1998.

\bibitem[Meuleau et~al.(2013)Meuleau, Kim, Kaelbling, and Cassandra]{MPK99}
N.~Meuleau, K.~Kim, L.~P. Kaelbling, and A.~R. Cassandra.
\newblock Solving {POMDPs by Searching the Space of Finite Policies}.
\newblock \emph{CoRR}, abs/1301.6720, 2013.
\newblock URL \url{http://arxiv.org/abs/1301.6720}.

\bibitem[Zhou and Hansen(2001)]{ZhH01}
R.~Zhou and E.~A. Hansen.
\newblock An {Improved Grid-Based Approximation Algorithm for POMDP}s.
\newblock In \emph{Proc. of the 17th International Joint Conf. on Artificial
  Intelligence}, page 707–714, San Francisco, CA, USA, 2001.
\newblock ISBN 1558608125.

\bibitem[Yu and Bertsekas(2012)]{YuB04}
H.~Yu and D.~Bertsekas.
\newblock Discretized {A}pproximations for {POMDP} with {A}verage {C}ost.
\newblock \emph{arXiv preprint arXiv:1207.4154}, 2012.

\bibitem[Baxter and Bartlett(2001)]{BaB01}
J.~Baxter and P.~L. Bartlett.
\newblock Infinite-{Horizon Policy-Gradient Estimation}.
\newblock \emph{J. of AI Res.}, 15:\penalty0 319--350, 2001.

\bibitem[Yu(2012)]{Yu05}
H.~Yu.
\newblock A {Function Approximation Approach to Estimation of Policy Gradient
  for POMDP with Structured Policies}.
\newblock \emph{arXiv preprint arXiv:1207.1421}, 2012.

\bibitem[Estanjini et~al.(2012)Estanjini, Li, and Paschalidis]{ELP12}
R.~M. Estanjini, K.~Li, and I.~C. Paschalidis.
\newblock A least squares temporal difference actor-critic algorithm with
  applications to warehouse management.
\newblock \emph{Naval Research Logistics}, 59:\penalty0 197--211, 2012.

\bibitem[Somani et~al.(2013)Somani, Ye, Hsu, and Lee]{despot}
A.~Somani, N.~Ye, D.~Hsu, and W.~S. Lee.
\newblock D{ESPOT: Online POMDP Planning with Regularization}.
\newblock In \emph{Advances in NeurIPS 26}, pages 1772--1780, 2013.

\bibitem[Mnih et~al.(2016)Mnih, Badia, Mirza, Graves, Lillicrap, Harley,
  Silver, and Kavukcuoglu]{A3C}
V.~Mnih, A.~P. Badia, M.~Mirza, A.~Graves, T.~Lillicrap, T.~Harley, D.~Silver,
  and K.~Kavukcuoglu.
\newblock Asynchronous {Methods for Deep Reinforcement Learning}.
\newblock In \emph{International conference on machine learning}, pages
  1928--1937, 2016.

\bibitem[Schulman et~al.(2017)Schulman, Wolski, Dhariwal, Radford, and
  Klimov]{PPO}
J.~Schulman, F.~Wolski, P.~Dhariwal, A.~Radford, and O.~Klimov.
\newblock Proximal {Policy Optimization Algorithms}.
\newblock \emph{arXiv preprint arXiv:1707.06347}, 2017.

\bibitem[Mnih et~al.(2013)Mnih, Kavukcuoglu, Silver, Graves, Antonoglou,
  Wierstra, and Riedmiller]{DQN_Rainbow}
V.~Mnih, K.~Kavukcuoglu, D.~Silver, A.~Graves, I.~Antonoglou, D.~Wierstra, and
  M.~Riedmiller.
\newblock Playing {Atari with Deep Reinforcement Learning}.
\newblock \emph{arXiv preprint arXiv:1312.5602}, 2013.

\bibitem[Capitan et~al.(2013)Capitan, Spaan, Merino, and
  Ollero]{capitan2013decentralized}
J.~Capitan, M.~T. Spaan, L.~Merino, and A.~Ollero.
\newblock Decentralized multi-robot cooperation with auctioned {POMDPs}.
\newblock \emph{The International Journal of Robotics Research}, 32\penalty0
  (6):\penalty0 650--671, 2013.

\bibitem[Foerster et~al.(2018)Foerster, Farquhar, Afouras, Nardelli, and
  Whiteson]{foerster2018counterfactual}
J.~N. Foerster, G.~Farquhar, T.~Afouras, N.~Nardelli, and S.~Whiteson.
\newblock Counterfactual {Multi-Agent Policy Gradients}.
\newblock In \emph{Thirty-second AAAI conference on artificial intelligence},
  2018.

\bibitem[{Bhattacharya} et~al.(2020){Bhattacharya}, {Badyal}, {Wheeler}, {Gil},
  and {Bertsekas}]{Bhattacharya2020RL_PAPI}
S.~{Bhattacharya}, S.~{Badyal}, T.~{Wheeler}, S.~{Gil}, and D.~{Bertsekas}.
\newblock Reinforcement {Learning for POMDP: Partitioned Rollout and Policy
  Iteration With Application to Autonomous Sequential Repair Problems}.
\newblock \emph{IEEE Robotics and Automation Letters}, 5\penalty0 (3):\penalty0
  3967--3974, 2020.

\bibitem[Yang et~al.(2018)Yang, Bellingham, Dupont, Fischer, Floridi, Full,
  Jacobstein, Kumar, McNutt, Merrifield, Nelson, Scassellati, Taddeo, Taylor,
  Veloso, Wang, and Wood]{grandChallengesWood}
G.-Z. Yang, J.~Bellingham, P.~E. Dupont, P.~Fischer, L.~Floridi, R.~Full,
  N.~Jacobstein, V.~Kumar, M.~McNutt, R.~Merrifield, B.~J. Nelson,
  B.~Scassellati, M.~Taddeo, R.~Taylor, M.~Veloso, Z.~L. Wang, and R.~Wood.
\newblock The grand challenges of {Science R}obotics.
\newblock \emph{Science Robotics}, 3, 2018.
\newblock \doi{10.1126/scirobotics.aar7650}.

\bibitem[{Waharte} and {Trigoni}(2010)]{SARuavs}
S.~{Waharte} and N.~{Trigoni}.
\newblock Supporting {Search and Rescue Operations with UAV}s.
\newblock In \emph{2010 International Conf. on Emerging Security Technologies},
  pages 142--147, 2010.

\bibitem[Cassandra(1998)]{Cassandra2003}
A.~Cassandra.
\newblock A {Survey of POMDP Applications}.
\newblock \emph{Working Notes of AAAI 1998 Fall Symposium on Planning with
  POMDP}, 1998.

\bibitem[Veres et~al.(2011)Veres, Molnar, Lincoln, and
  Morice]{decisionMakingAutonomousVeh}
S.~M. Veres, L.~Molnar, N.~K. Lincoln, and C.~P. Morice.
\newblock Autonomous vehicle control systems - a review of decision making.
\newblock \emph{Proc. of the Institution of Mechanical Engineers, Part I:
  Journal of Systems and Control Engineering}, 225:\penalty0 155--195, 2011.

\bibitem[Dunbabin et~al.(2009)Dunbabin, Corke, Vasilescu, and
  Rus]{controlRusUnderwater}
M.~Dunbabin, P.~Corke, I.~Vasilescu, and D.~Rus.
\newblock Experiments with {Cooperative Control of Underwater R}obots.
\newblock \emph{IJRR}, 28:\penalty0 815--833, 2009.

\bibitem[Das et~al.(2015)Das, Py, Harvey, Ryan, Gellene, Graham, Caron, Rajan,
  and Sukhatme]{serviceRobotGaurav}
J.~Das, F.~Py, J.~B. Harvey, J.~P. Ryan, A.~Gellene, R.~Graham, D.~A. Caron,
  K.~Rajan, and G.~S. Sukhatme.
\newblock Data-driven robotic sampling for marine ecosystem monitoring.
\newblock \emph{IJRR}, 34:\penalty0 1435--1452, 2015.

\bibitem[Schwarting et~al.(2018)Schwarting, Alonso-Mora, and
  Rus]{decisionMakingRus}
W.~Schwarting, J.~Alonso-Mora, and D.~Rus.
\newblock Planning and {D}ecision-{M}aking for {A}utonomous {V}ehicles.
\newblock \emph{Annual Review of Control, Robotics, and Autonomous Systems},
  1:\penalty0 187--210, 2018.

\bibitem[Everett et~al.(2018)Everett, Chen, and How]{AIHowDeepRL}
M.~Everett, Y.~F. Chen, and J.~P. How.
\newblock Motion {Planning Among Dynamic, Decision-Making Agents with Deep
  Reinforcement Learning}.
\newblock In \emph{2018 IEEE/RSJ International Conf. IROS}, pages 3052--3059,
  2018.

\bibitem[Kretzschmar et~al.(2016)Kretzschmar, Spies, Sprunk, and
  Burgard]{RLWolfram}
H.~Kretzschmar, M.~Spies, C.~Sprunk, and W.~Burgard.
\newblock Socially {Compliant Mobile Robot Navigation via Inverse Reinforcement
  Learning}.
\newblock \emph{IJRR}, 35:\penalty0 1289--1307, 2016.

\bibitem[S{\"u}nderhauf et~al.(2018)S{\"u}nderhauf, Brock, Scheirer, Hadsell,
  Fox, Leitner, Upcroft, Abbeel, Burgard, Milford, et~al.]{deepLearningAIAbeel}
N.~S{\"u}nderhauf, O.~Brock, W.~Scheirer, R.~Hadsell, D.~Fox, J.~Leitner,
  B.~Upcroft, P.~Abbeel, W.~Burgard, M.~Milford, et~al.
\newblock The limits and potentials of deep learning for robotics.
\newblock \emph{IJRR}, 37:\penalty0 405--420, 2018.

\bibitem[Ingrand and Ghallab(2017)]{AIDeliberation}
F.~Ingrand and M.~Ghallab.
\newblock Deliberation for autonomous robots: {A survey}.
\newblock \emph{Artificial Intelligence}, 247:\penalty0 10--44, 2017.

\bibitem[Omidshafiei et~al.(2015)Omidshafiei, Agha-Mohammadi, Amato, and
  How]{POMDPAmatoDecentralized}
S.~Omidshafiei, A.-A. Agha-Mohammadi, C.~Amato, and J.~P. How.
\newblock Decentralized control of {Partially Observable Markov Decision
  Processes using belief space macro-actions}.
\newblock In \emph{IEEE ICRA}, pages 5962--5969, Seattle, WA, 2015.

\bibitem[Yi et~al.(2019)Yi, Nam, and Sycara]{POMDPSycara}
S.~Yi, C.~Nam, and K.~Sycara.
\newblock Indoor {Pursuit-Evasion with Hybrid Hierarchical Partially Observable
  Markov Decision Processes for Multi-robot S}ystems.
\newblock In \emph{Distributed Autonomous Robotic Systems}, volume~9, pages
  251--264, 2019.

\bibitem[Gil et~al.(2015)Gil, Kumar, Katabi, and Rus]{gilMultiRobot}
S.~Gil, S.~Kumar, D.~Katabi, and D.~Rus.
\newblock Adaptive {Communication in Multi-Robot Systems Using Directionality
  of Signal Strength}.
\newblock \emph{IJRR}, 34:\penalty0 946--968, 2015.

\bibitem[Wang et~al.(2019)Wang, Jadhav, Vohs, Hughes, Mazumder, and
  Gil]{gilISRR2019}
W.~Wang, N.~Jadhav, P.~Vohs, N.~Hughes, M.~Mazumder, and S.~Gil.
\newblock Active {Rendezvous for Multi-Robot Pose Graph Optimization using
  Sensing over Wi-Fi}.
\newblock \emph{arXiv preprint arXiv:1907.05538}, 2019.

\bibitem[Gil et~al.(2012)Gil, Feldman, and Rus]{gilMultiRobotCov}
S.~Gil, D.~Feldman, and D.~Rus.
\newblock Communication {Coverage for Independently Moving Robots}.
\newblock In \emph{2012 IEEE/RSJ International Conf. on IROS}, pages
  4865--4872, Vilamoura, 2012.

\bibitem[Lagoudakis and Parr(2003)]{LaP03}
M.~G. Lagoudakis and R.~Parr.
\newblock Reinforcement {Learning as Classification: Leveraging Modern
  Classifiers}.
\newblock In \emph{Proc. 20th ICML}, pages 424--431, 2003.

\bibitem[Dimitrakakis and Lagoudakis(2008)]{DiL08}
C.~Dimitrakakis and M.~G. Lagoudakis.
\newblock Rollout {Sampling Approximate Policy Iteration}.
\newblock \emph{Mach. Learn.}, 72:\penalty0 157--171, 2008.

\bibitem[Yemini et~al.(2020)Yemini, Gil, and Goldsmith]{globecomm2020}
M.~Yemini, S.~Gil, and A.~Goldsmith.
\newblock Exploiting local and cloud sensor fusion in intermittently connected
  sensor networks.
\newblock \emph{arXiv preprint arXiv:2005.12495}, 2020.

\end{thebibliography}

\clearpage
\appendix
\section{Appendix: Additional Computational Results}
\subsection{\SG{Imperfect Communication of Controls}}
\label{sec:communication_less}
In this appendix we discuss computational results with schemes where the agents do not fully share their information. Instead, they use estimates of other agents' controls (what we refer to as ``signaling'') that can be precomputed, combined with belief state estimation.  By doing this, the agents can obtain greater computational speedup through parallelization.

\SG{We consider the case where agents cannot communicate their choice of control with one another, although we assume that the belief state is still shared.  This case may arise when agents have access to information that changes more slowly, such as beliefs, but cannot necessarily share information at tighter timescales such as chosen controls. In this context, we consider several different communication architectures and study their effect on the resulting performance of our multiagent rollout methods.}
\vspace{-5pt}
\paragraph{a) Approximate multiagent rollout with base policy signaling and shared belief (\acrshort{amr_b}):}

\SG{
Here, $m$ independent minimizations are performed, once over each of the agent controls $\bar{u}_1,\ldots,\bar{u}_m$, with all other controls estimated by the base policy.} 
\SB{In other words, the control component for agent $\ell$ is determined as follows.
\begin{equation}
\label{eq:amr_shared_belief}
\bar{u}_\ell \in \arg\min_{u_\ell\in U_\ell}\hat g(b,u') + \alpha \sum_{z\in Z}\hat p(z\,|\, b,u') J_{\mu}\big(F(b,u',z)\big) 
\end{equation}
where $u'=(u_{1:\ell-1},u_\ell,u_{\ell+1:m})$ 
and $u_k$ generically denotes the base policy’s control component for agent $k$. This is the method where the agents do not communicate their controls to each other. Note that as discussed earlier, the performance of this method may strongly depend on the initial locations of the agents. In particular, if all agents start at the same location, the performance of this method can be very poor and may result in oscillatory agent motions. On the other hand, other initial agent locations may result in much better performance for this method. This is reflected in the results of Table~\ref{table:AMR_shared_belief}, which give the ``aggregate” performance of the method (averaged over many randomly chosen initial agent positions).
This represents the extreme case of no communicated controls.}
\vspace{-5pt}
\paragraph{b) Approximate multiagent rollout with neural network policy signaling and shared belief (\acrshort{amr_n}):}\SG{
In this approach, $m$ independent minimizations are performed, once over each agent controls $\bar{u}_1,\ldots,\bar{u}_m$, with predecessors' control components predicted by a neural network that approximates the one-agent-at-a-time rollout policy, and successors' control components estimated by the base policy.}
\SB{The control component of agent $\ell$ is determined by Eq. (\ref{eq:amr_shared_belief}) where $u'=(\hat{u}_{1:\ell-1},u_\ell,u_{\ell+1:m})$ and $u_k$ is the base policy's control component for agent $k$ and $\hat{u}_k$ is the neural network policy's control component for agent $k$.
This also falls into the extreme case of no communicated controls.}
\vspace{-5pt}
\paragraph{c) Approximate multiagent rollout with best PI policy signaling and shared belief (\acrshort{amr_pi}):}\SB{
This approach is similar to \acrshort{amr_n}. 
Instead of using the neural network policy that approximates the one-agent-at-a-time rollout, the predecessors' control components are given by the neural network corresponding to one of the approximate policy iterations (possibly the best iteration) and the base policy is given by the previous policy iteration.}
\vspace{-5pt}
\paragraph{d) Approximate multiagent rollout with local communication and shared belief (\acrshort{amr_lc}):}\SG{
In this approach, we consider a local communication scheme where the computed predecessors' control components are communicated among agents when the corresponding agents are less than a radius of $r$ hops away on the graph, and all other controls are estimated by the base policy. }
\SB{The control component of agent $\ell$ is determined by Eq. (\ref{eq:amr_shared_belief}) where $u'=(\bar{\bar{u}}_{1:\ell-1},u_\ell,u_{\ell+1:m})$ and $u_k$ is the base policy's control component for agent $k$. $\bar{\bar{u}}_k$ is the one-agent-at-a-time rollout policy's $k^{th}$ control component ($\tilde{u}_k$) if agent $k$ and agent $\ell$ are within $r$ hops away, otherwise it is set to $u_k$.}
\vspace{-5pt}
\paragraph{e) Approximate multiagent rollout with intermittent as well as local communication and shared belief (\acrshort{amr_ilc}):}\SG{
In this approach, we assume intermittent connectivity to a centralized cloud server that provides access to computed predecessors' control components with probability $\rho>0$. With a probability $1-\rho$, the method assumes local communication with a radius of $r$. In other words, when the cloud is available, the one-agent-at-a-time rollout is performed, and otherwise, the method follows AMR-LC, as described earlier.
This architecture strikes a practical tradeoff since it takes advantage of a rich centralized information source whenever possible and uses clustered local communication when the server is unreachable.}

\SG{Table~\ref{table:AMR_shared_belief} presents numerical results demonstrating the performance of these different architectures involving imperfect control communication (but perfect belief sharing), and compares them with our one-agent-at-a-time rollout with perfect communication for 4,8,10 agents, respectively.} 
\SG{We observe that AMR-B gives worse performance than other multiagent rollout methods. This is because it performs local optimizations for each agent without any coordination, i.e., no communicated controls. 
\Doubt{This method is also susceptible to oscillations. In an initial scenario where both agents start from the same location which is almost equidistant from the nearby damaged locations, each agent thinks that the other agents will fix the nearest damaged location and goes in the opposite direction and, in effect, oscillates indefinitely}.
AMR-N performs better since it uses a neural network to approximate the one-agent-at-a-time rollout policy, which is then used to estimate predecessor agent controls. The performance of this approach improves with more accurate policy networks. AMR-PI performs better than AMR-N which can be explained by its usage of better base and signaling policies. AMR-LC works very well for our problem since the spatial clustering of agents makes sense in this network repair context. However, this approach is heavily dependent on the communication radius $r$ (we use $r=2$ in our experiments) and can exhibit poor behavior if coordination is required across agent clusters (i.e., the damage is in a distant part of the environment). This variant is problem dependent and assumes agents can always communicate controls perfectly with other agents within $r$ hops. This may not be a practical assumption in other multiagent POMDP problems. AMR-ILC performs best among all other approximate multiagent rollout approaches by utilizing all past controls whenever available by means of accessing the centralized cloud. Naturally, the cost of generated policy improves with better connection probability $\rho$.
This suggests that our multiagent rollout methods can produce intelligent policies starting from a simple base policy even with an imperfect and/or intermittent communication setting.}

	\begin{table}[hbt!]
    \centering
    \caption{\SG{Cost comparison of base, standard rollout ($4$ agents only), one-agent-at-a-time rollout, and different approximate multiagent rollout policies involving imperfect control communication (assuming a shared belief)}}
    \label{table:AMR_shared_belief}
    \begin{tabular}{|p{4mm}|p{5mm}|p{7mm}|p{6mm}|p{9mm}|p{9mm}|p{9mm}|p{10mm}|p{11mm}|p{11mm}|p{11mm}|}
    \hline
    \scriptsize
    agent&\scriptsize base&\scriptsize standard rollout & \scriptsize 1-at-a-time &\scriptsize AMR-B &\scriptsize AMR-N &\scriptsize AMR-PI&\scriptsize{AMR-LC $r$=2} &\scriptsize AMR-ILC $\rho$=0.8, $r$=2 &\scriptsize AMR-ILC $\rho$=0.5, $r$=2 &\scriptsize AMR-ILC $\rho$=0.3, $r$=2\\
        \hline
         4 &3277  &1879& 1925 & 3187 &2635 &-& 2038 & 1946 & 1964 & 1976\\ \hline
         8 &5347 &- & 992 & 2513 &1712 & 1590&1010 & 992 & 998 & 1005\\ \hline
         10 & 4667 &- & 799 & 2487 &1533& 1428 & 813 & 804 & 807 & 809\\
         \hline
    \end{tabular}
\end{table}

\subsection{\SB{Imperfect Communication of Belief States and Controls}}
\label{sec:communication}
\vspace{-5pt}
\SB{Here we consider the case where the agents do not share their belief states and cannot communicate their choice of control with one another at all times. Each agent knows its location and can obtain a perfect observation of the damage at its location. However, agents may not always perfectly perceive other agents' locations, have knowledge of their observations. In this way, each agent may have a local belief state that is different from the true global belief state. We consider the existence of a centralized cloud server having access to the global belief states with an intermittent connection probability of $\rho\in(0,1)$ (a hybrid communication infrastructure similar to the one analyzed in~\cite{globecomm2020}). If the cloud is reachable, every agent can access the true belief state and the computed predecessors' control components. In this context, we consider several different communication architectures and perform an extensive performance simulation study.}
\vspace{-5pt}
\paragraph{a) Approximate multiagent rollout with intermittent communication and base policy signaling (\acrshort{amr_ib1}):}
\SB{
If the cloud is not accessible, each agent performs one optimization to estimate its own control component evaluated at its local belief state assuming other agents' control components are given by the base policy. The local belief corresponding to an agent evolves by applying the locally computed control component and the base policy's control components for other agents. When the cloud is accessible, agents synchronize with the global belief state that gives the most up-to-date damage distributions, locations, and observations of all agents. During that time step, the one-agent-at-a-time rollout is performed with the computed predecessors' control components given by the cloud, and each agents' belief state is evolved forward using this information accordingly.
}
\vspace{-5pt}
\paragraph{b) Approximate multiagent rollout with intermittent communication and base policy (\acrshort{amr_ib0}):}
\SB{If the cloud is not accessible, each agent chooses to execute the base policy control evaluated at its local belief state independently from the team (i.e. without estimating or taking into consideration the actions of the other agents). For propagating forward the belief, each agent assumes that all other agents are also choosing actions using the base policy. When the cloud is accessible, agents synchronize with the global belief state, and the one-agent-at-a-time rollout is performed with the computed predecessors' control components given by the cloud. Each agent's belief state evolves by applying this rollout control.}

\SB{Table \ref{table:AMR_1} shows the cost comparison between the base policy, one-agent-at-a-time rollout policy, and
different architectures involving intermittent communication with imperfect belief and control sharing
for $4,8$, and $10$ agents respectively. 
We observe that all the multiagent rollout methods under the intermittent communication assumption improve over the base policy, and the cost improves with a better connection probability $\rho$. With a high probability of $\rho \rightarrow 1$, the methods perform similar to the one-agent-at-a-time rollout, attaining the same behavior as one-agent-at-a-time rollout when $\rho=1$. With a low probability of $\rho \rightarrow 0$, we observe that the methods produce similar costs to the base policy. Interestingly, we see that for the same intermittent communication probability $\rho$, \SB{AMR-IB0 outperforms AMR-IB1 in some cases}, and at the same time, reduces the computations by a factor of $m$ when the cloud is not reachable. In AMR-IB1, each agent chooses its control component, thinking the other agents will apply the base policy. This, in effect, might miss some damage locations before the global belief is shared. In contrast, AMR-IB0 does not try to make any smarter moves until the cloud is reachable and uses the base policy's controls until then. This method has less chance of missing some damage location than AMR-IB1, and gives better cost. }

\SB{
The simulation results suggest that our multiagent rollout algorithm and its variants are suitable for a practical class of multiagent systems involving intermittent communication and imperfect state observation. Our methods take advantage of a centralized information source whenever possible and use it to select controls with imperfect knowledge of the true global belief state when the server is unreachable.}
	
%
\begin{table}[hbt!]
    \centering
    \caption{\SB{Cost comparison of base, one-agent-at-a-time rollout, and approximate multiagent rollout policies with different intermittent communication architectures and connection probabilities ($\rho$)}}
    \label{table:AMR_1}
    
    \begin{tabular}{|p{7mm}|p{10mm}|p{12mm}|p{20mm}|p{20mm}|p{20mm}|p{20mm}|}
    \hline
    \scriptsize agents&\scriptsize base&\scriptsize {1-at-a-time}&\scriptsize{AMR-IB1 $\rho=0.8$} &\scriptsize {AMR-IB0 $\rho=0.8$}&\scriptsize {AMR-IB1 $\rho=0.4$} &\scriptsize{AMR-IB0 $\rho=0.4$} \\
        \hline
         4 &3277  & 1925 & 2303.96 &2239.67 & 2793.45 &2767.4 \\ \hline
         8 &5347 & 992 & 1127.66 & 1140.21& 1512.79&1713.17 \\ \hline
         10 & 4667 & 799 &960.104 & 920.58 & 1265.49 &1398.94 \\
         \hline
    \end{tabular}
\end{table}

\clearpage

\printglossary[type=\acronymtype]
\printglossary

Related video lecture about our multiagent rollout methods can be found:
https://www.youtube.com/watch?v=eqbb6vVlN38\&t=302s

\end{document}